\documentclass[journal, onecolumn, draftcls]{IEEEtran}

\usepackage{algorithm2e}
\usepackage{graphicx}
\usepackage{epsfig}
\usepackage{subfigure}
\usepackage{latexsym}
\usepackage{amsmath}
\usepackage{amssymb}
\usepackage{multirow}
\include{new_commands}

\def\N{\hbox{I \kern-.5em N}}
\def\M{\hbox{I \kern-.5em M}}
\def\J{\hbox{I \kern-.5em J}}
\def\B{\hbox{I \kern-.5em B}}
\def\Q{\hbox{I \kern-.5em Q}}
\def\D{\hbox{I \kern-.5em D}}
\def\G{\hbox{I \kern-.5em G}}
\def\E{\hbox{I \kern-.5em E}}
\def\F{\hbox{I \kern-.5em F}}
\def\A{\hbox{I \kern-.5em A}}
\def\H{\hbox{I \kern-.5em H}}
\def\F{\hbox{I \kern-.5em F}}
\def\L{\hbox{I \kern-.5em L}}
\def\P{\hbox{I \kern-.5em P}}
\def\X{\hbox{I \kern-.5em X}}
\def\R{\hbox{I \kern-.5em R}}

\begin{document}

\title{Low-Complexity Particle Swarm Optimization for Time-Critical Applications}

%\author{Muhammad S. Sohail, Muhammad O. Bin Saeed, Syed Z. Rizvi, \\
%Mubien Shoaib and Asrar U. H. Sheikh}

\author{Muhammad~S.~Sohail, Muhammad~O.~Bin~Saeed,~\IEEEmembership{Member}, Syed~Z.~Rizvi,~\IEEEmembership{Student Member} \\ Mobien Shoaib,~\IEEEmembership{Member} and Asrar U. H. Sheikh,~\IEEEmembership{Fellow}
%\thanks{M. Shell is with the Department
%of Electrical and Computer Engineering, Georgia Institute of Technology, Atlanta,
%GA, 30332 USA e-mail: (see http://www.michaelshell.org/contact.html).}% <-this % stops a space
%\thanks{J. Doe and J. Doe are with Anonymous University.}% <-this % stops a space
%\thanks{Manuscript received April 19, 2005; revised January 11, 2007.}}
}

\date{}
\maketitle

\begin{abstract}
Particle swam optimization (PSO) is a popular stochastic optimization method that has found wide applications in diverse fields. However, PSO suffers from high computational complexity and slow convergence speed. High computational complexity hinders its use in applications that have limited power resources while slow convergence speed makes it unsuitable for time critical applications. In this paper, we propose two techniques to overcome these limitations. The first technique reduces the computational complexity of PSO while the second technique speeds up its convergence. These techniques can be applied, either separately or in conjunction, to any existing PSO variant. The proposed techniques are robust to the number of dimensions of the optimization problem. Simulation results are presented for the proposed techniques applied to the standard PSO as well as to several PSO variants. The results show that the use of both these techniques in conjunction results in a reduction in the number of computations required as well as faster convergence speed while maintaining an acceptable error performance for time-critical applications.
\end{abstract}


%%%%%%%%%%%%%%%%%%%%%%%%%%%%%%%%%%%%%%%%%%%%%%%%%%%%%%%%%%%%
%           Section I
%%%%%%%%%%%%%%%%%%%%%%%%%%%%%%%%%%%%%%%%%%%%%%%%%%%%%%%%%%%%

\section{Introduction}
\label{Sec:Intro}
\label{intro}

Since its introduction, the particle swarm optimization (PSO) algorithm \cite{Eberhart95}-\cite{Kennedy95} has attracted a lot of attention due to its ability to effectively solve a large number of problems in diverse fields. Optimization problems generally involve maximization or minimization of an objective function, often subject to multiple constraints. The notion of using a swarm of agents that help each other in a directed search to find an optimal solution to a problem has been received with much enthusiasm by the research community globally, as it brought a much-needed enrichment to the existing computational arsenal. % Since then, mathematicians and scientists have strived to improve the performance of the basic PSO algorithm in several ways. Some of the major advancements in this regard are discussed below.

The improvements and advancements in the PSO algorithm resulted in algorithms that are used in numerous applications.
%PSO has been applied on a variety of problems.
For example, the authors in \cite{DiCarro2005} used PSO and ant colony algorithms to devise an adaptive routing algorithm for a mobile ad hoc network. The work in \cite{Yuan2006} used the PSO algorithm to propose a multihop virtual MIMO communication protocol with a cross-layer design to improve the performance and life of a wireless sensor network. PSO-based adaptive channel equalization was proposed in \cite{Awami2007}-\cite{Yogi2010a} while \cite{Soo2007}-\cite{Choudhry2010} used PSO for multiuser detection in CDMA systems. The works in \cite{Jouny2008}-\cite{Zeng2011} used the PSO algorithm for radar design. PSO was used for MIMO receivers \cite{Xiao2008}, adaptive beamforming \cite{Zaharis2011} and economic dispatch problems \cite{Gaing2003}-\cite{Chaturverdi2008}. PSO also found use in several antenna applications \cite{Jin2007} such as the design of phased arrays \cite{Boeringer2004}-\cite{Khodier2005} and broadband antenna design \cite{Jin2005}. Other applications in which the PSO algorithm was successfully implemented include controller design \cite{Gaing2004}-\cite{Yoshida2000}, power systems \cite{Abido2002}, biometrics \cite{Veera2005}, distribution networks \cite{Kannan2005}, electromagnetics \cite{Robinson2004}, robotics \cite{Chatterjee2005}, gaming \cite{Messerschmidt2004} and data clustering \cite{Chen2004}, \cite{Yuwono2012}, \cite{Merwe2003}. A detailed survey of PSO applications can be found in \cite{Poli2007}-\cite{Poli2008}.

Despite the popularity of PSO, it suffers from the issues of high computational complexity and considerable convergence time. This hinders its use in applications that require fast convergence or have power/computational constraints.

In this work, we propose two new techniques aimed at reducing computational complexity as well as improving the speed of convergence of the PSO algorithm. An event-triggering-based approach reduces the number of computations for an acceptable degradation in performance. This is followed by a dimension-wise cost function evaluation technique that increases the speed of convergence as well as improving overall performance. The techniques can be applied to any existing PSO variant either exclusively, or in conjunction.

The paper is arranged as follows. In Section \ref{LitSurvey}, we take a look at some of the important variants of PSO and assess in detail the merits and costs of each algorithm. In Section \ref{Event-Trig} we present an event-triggering-based approach to reduce the computational complexity of the particle swarm search algorithm. Section \ref{Dim-Wis} takes a look at the dimension-wise search for particle position, aimed at obtaining faster convergence towards optimal solutions. Section \ref{combo} discusses how the two approaches can be combined for optimum results. Section \ref{Results} discusses the experimental results while Section \ref{Conclusion} presents the conclusion.

\section{The particle swarm optimizer and its variants}
\label{LitSurvey}

We start with a brief recap of the working of the PSO algorithm and then discuss some of its variants. %While it is assumed that the reader already possesses a basic understanding of the working of the particle swarm search, details of the original algorithm are presented very briefly here.
The PSO algorithm is a population based optimization method that imitates swarm behavior observed in a herd of animals or a school of fish. The swarm consists of particles representing the set of optimization parameters. Each particle searches for a potential solution in a multi-dimensional search space. The aim of the algorithm is to converge to an optimum value for each parameter. The estimate for each particle is tested using a fitness function. The fitness function when evaluated, gives a fitness value that defines the "goodness" of the estimate.
%The aim of the algorithm is to converge to an optimum value for each parameter. The estimate for each particle is tested using a certain fitness value that defines the ``goodness'' of the estimate.
Each particle has two state variables, namely ``particle position'' $x(i)$ and ``particle velocity'' $v(i)$ , where $i$ denotes iteration index. The fitness of an individual particle is measured by evaluating the cost function for that particle. Every particle combines its own best attained solution with the best solution of the whole swarm to adapt its search pattern. For a swarm of $N$ particles traversing a $D$-dimensional space, the velocity and position of each particle are updated as \cite{Eberhart95}
\begin{eqnarray}
%\begin{align}
\label{v_orig}
v_{k}^{d} \left(i+1\right) =&& {\hspace{-1.7em}}v_{k}^{d} \left(i\right) + c_1 \cdot r_{1,k} (i) \cdot \left(p_{k}^{d} - x_{k}^{d} \left(i\right) \right)\nonumber \\ && {\hspace{0.5em}} + c_2 \cdot r_{2,k} (i) \cdot \left(g^{d} - x_{k}^{d} \left(i\right) \right),\\
x_{k}^{d}\left(i+1\right) =&& {\hspace{-1.7em}}x_{k}^{d}\left(i\right) + v_{k}^{d}\left(i+1\right), \label{x_orig}
%\end{align}
\end{eqnarray}
where $d = 1,\cdots,D$ denotes the particle's dimension index and $k = 1,\cdots,N$ is the particle index. The constants $c_1$ and $c_2$ are called the cognitive and social parameters, and are responsible for defining a particle's affinity towards its own best and the global best solutions, respectively. The variables $v_{k}^{d}$ and $x_{k}^{d}$ are the velocity and position of the $k$-th particle corresponding to its $d$-th dimension while $p_{k}^{d}$ and $g^{d}$ are the particle's local best and the swarm's global best positions for the $d$-th dimension respectively. The variables $r_{1,k}$ and $r_{2,k}$ are drawn from a uniform random distribution $[0,1]$ and are the source of randomness in the search behavior of the swarm.

Of the earliest variants to the original algorithm, Eberhart and Shi proposed an inertia-weight model \cite{Shi98}, which multiplies the velocity of the current iteration with a factor known as the inertia weight
\begin{eqnarray}
%\begin{align}
\label{v_inertia}
v_{k}^{d} \left(i+1\right) = && {\hspace{-1.7em}} w \cdot v_{k}^{d} \left(i\right) + c_1 \cdot r_{1,k} (i) \cdot \left(p_{k}^{d} - x_{k}^{d} \left(i\right) \right)\nonumber \\ && {\hspace{0.6cm}}+  c_2 \cdot r_{2,k} (i) \cdot \left(g^{d} - x_{k}^{d} \left(i\right) \right).
%\end{align}
\end{eqnarray}

The inertia weight, $\;w \; \in \; [0,1]$, ensures convergence and controls the momentum of the particle. If $w$ is too small, very little momentum is preserved from the previous iteration and thus quick changes of direction are possible. If $w = 0$, the particle moves without knowledge of the past velocity profile. On the other hand, a very high value of inertia weight means that the particles can hardly change their direction, which translates into more exploration and slow convergence. Maurice and Clerc later introduced a constriction-factor \cite{Clerc02} to constrict the velocity by multiplying it with this factor before updating the position of the particle. The constriction factor model has the same effect as the inertia weight model, except for the fact that it scales the contributions from both the local and global best solutions, thus limiting the search space. Low values facilitate rapid convergence but at the cost of smaller exploration space and vice versa.

This initial model of the PSO optimizer is very efficient in converging to near-optimal solutions for uni-modal problems, i.e. problems that do not have local minima. The algorithm therefore %does not get entrapped into a local minimum and therefore
converges quickly to optimal or near-optimal solution. However, when the problem is multi-modal, then too fast a convergence often leads to entrapment of the algorithm in a local optimum. The algorithm must explore different solution areas in such a case before converging in order to avoid getting stuck in a sub-optimal region.

Shortly after the first few PSO variants appeared in literature, significant progress was made towards solving multi-modal problems with the development of neighborhood topologies. It was suggested that instead of each particle being connected to the whole swarm, it should be connected to a small neighborhood of particles and if each neighborhood had its own global best solution, then different solution areas of the swarm would be explored, thus avoiding entrapment in local minima. Thus, in the local version of PSO (LPSO), the velocity of each particle is adjusted according to its personal best and the best performance achieved so far within its neighborhood. Kennedy and Mendes have discussed the effects of various population topologies on the PSO algorithm in \cite{Kennedy02}. To test this, benchmark problems were solved using different topologies. The authors stated that information moves fastest between connected pairs of particles, and is slowed down by the presence of intermediate particles. They further concluded that the conventional PSO topology known as ``global best topology'' facilitates the most immediate communication possible. However, on complex multi-modal problems, this is not necessarily desirable and the population will fail to explore outside of locally optimal regions.

The ring topology known as ``local best'' is the slowest and most indirect communication pattern and a solution found by a particle moves very slowly around the ring. However, this ensures proper exploration which is desirable in multi-modal complex problems. One variant, proposed by Peram et. al., is known as fitness-distance-ratio (FDR) PSO, with near neighbor interactions \cite{Peram03}. The FDR-PSO selects one other particle having higher fitness value near the particle being updated, in the velocity updating equation. Other variants which used multi-swarm \cite{Blackwell04} or sub-population \cite{Lovbjerg01} are also generally included under LPSOs since the sub-groups are treated as special neighborhood structures.

Similar conclusions were derived by Liang and Suganthan who proposed a dynamic multi-swarm particle swarm (DMS-PSO) \cite{Liang05}. It can be said that in topology-based LPSO algorithms, a trade-off between the speed of exploration and depth of exploration hinges upon the ``granularity'' of the neighborhood used. Liang and Suganthan showed with simulations that on uni-modal benchmark functions like the sphere function, all algorithms perform almost equally well. %However, the DMS-PSO performs the best on the multi-modal problems, enforcing the same conclusion that local versions work better on multi-modal problems.

Mendes et al. proposed a fully informed particle swarm (FIPS) \cite{Mendes04}, in which all the neighborhood is a source of influence rather than just one particle. Chatterjee and Siarry \cite{Chatterjee06} presented a modified version of the PSO algorithm where they proposed a nonlinear variation of inertia weight along with a particle's old velocity in order to speed up the convergence as well as fine tune the search in the multi-dimensional space. Hu and Eberhart \cite{Hu02} used a dynamic neighborhood where closest particles in the search space are selected to be a particle's new neighborhood in each generation. Parsopoulos and Vrahatis combined the global version and local version together and constructed a unified particle swarm optimizer UPSO \cite{Parsopoulos04}.

There have been several efforts in investigating hybridization by combining PSO with other search techniques to improve the performance of the PSO. Evolutionary operators such as selection, crossover, and mutation were used in PSO to keep the best particles \cite{Angeline98}, to increase the diversity of the population, and avoid entrapment in a local minimum \cite{Lovbjerg01}. The use of mutation operators was explored to mutate parameters like inertia weight \cite{Miranda02}. Relocation of particles was used in \cite{Lovbjerg02}. In \cite{Blackwell02} collision-avoiding mechanisms were introduced to prevent particles from moving too close to each other. This, according to the authors, helps maintain the diversity in the swarm and helps to escape local optima. In \cite{Parsopoulos04b}, deflection, stretching, and repulsion techniques were used to find maximum possible minima by preventing particles from moving to a previously discovered minimal region. An orthogonal learning technique combined with orthogonal experimental design were used to enhance performance in orthogonal PSO (OPSO) \cite{Ho08} and orthogonal learning PSO (OLPSO) \cite{Zhan11}, although at the cost of very high computational complexity.

A cooperative particle swarm optimizer (CPSO) was proposed in \cite{Bergh04}. The CPSO uses one-dimensional swarms and searches each of them separately, so as to avoid the ``curse of dimensionality'' that plagues most stochastic optimization algorithms including PSO and genetic algorithms. These separate searches are then integrated together by a global swarm. This results in significant improvement over the performance of the standard PSO algorithm when solving multimodal problems albeit at a much higher computational cost.

Ratnaweera et al. suggested a self-organizing hierarchical PSO (HPSO) algorithm with time-varying acceleration coefficients \cite{Ratnaweera04}. The inertia weight term is removed with the idea that only the acceleration coefficients should guide the movement of the particle towards the optimum solution. Both the acceleration coefficients vary linearly with time. If the velocity goes to zero at some point, the particle is re-initialized using a predefined starting velocity. Due to this self-organizing and self-restarting property, the HPSO algorithm achieves outstanding results.

The comprehensive learning PSO (CLPSO) algorithm of \cite{Liang06} provides the best performance-complexity trade-off among all existing PSO variants. It divides the unknown estimate into sub-vectors. A particle chooses two random neighbor particles for each sub-vector, and chooses a sub-vector as an exemplar based on the best fitness value for that particular sub-vector. The combined result of all sub-vectors gives the overall best vector, which is then used to perform the update. If the particle stops improving for a certain number of iterations, then the neighbor particles for the sub-vectors are changed. %The velocity update equation for the CLPSO algorithm is given by \cite{Liang06}
%where the bold letter indicates a vector parameter.
The success of this technique may be attributed to its exploitation of the ``best information'' profile (or trend) of all other particles in updating the velocity equation, as this rich historical information will help improve the predictive power of the velocity equation. The authors claimed that although their variant did not perform well on uni-modal problems, it outdid most variants on multi-modal problems. When solving real-world problems, one usually does not know the shape of the fitness landscape. Hence, the authors concluded that, in such cases, it is advisable to use an algorithm that performs well on multimodal problems.

However, it is our humble opinion that %HPSO and CLPSO,
the algorithms presented in literature, despite their accurate performances, are still slow in convergence and far computationally complex for time-critical and power-limited applications. Keeping in mind all of these very important works, our aim in this paper is to propose an algorithm that emphasizes on quick convergence and low-complexity while still maintaining significant accuracy acceptable to real world problems.

\section{Event-triggering in particle swarm optimization}
\label{Event-Trig}

The first technique is aimed at reducing the number of computations of the PSO algorithm. We begin by first analyzing the complexity of the standard PSO algorithm and then %motivate our approach.
we derive a motivation for our approach.

\subsection{Complexity of the Standard PSO Algorithm}

The number of computations required for a complete run of the PSO algorithm are the sum of the computations required to calculate the cost of a candidate solution (based on current position of the particles) and the computations required to update each particle's position (\ref{x_orig}) and velocity (\ref{v_inertia}). Both of these are directly proportional to the number of iterations.

The computational complexity of evaluating the cost function depends on the particular cost function under consideration. For example, the sphere function (see Table \ref{TestTable}) requires $D$ multiplications per particle per iteration, resulting in a total of $DN$ multiplications for cost function evaluation per iteration. Similarly, the Rosenbrock function (see Table \ref{TestTable}) requires $4DN$ multiplications. These computations need to be performed at every iteration for all PSO variants and cannot be reduced.

For the second set of computations (i.e., the ones required for the update equations), the standard PSO algorithm requires $5DN$ multiplications per iteration. This number is $5$ times the number of multiplications required for the sphere function and $1.25$ times that required by the Rosenbrock function. This shows that the cost associated with the update equations makes a significant contribution to the total computational cost of the PSO algorithm.

\subsection{Motivation}

%The PSO algorithm and all its variants try to reach the global minimum with the help of different approaches. All PSO variants randomly search the problem space before converging to a local or global minimum. The ``better'' versions tend to move out of the local minima in order to reach the global minimum. However, once a particle settles in a valley, it shifts from searching mode to minimization mode, i.e., it tries to minimize the error as much as possible. Here lies our motivation for the proposed technique.

It is observed that many PSO variants tend to achieve accuracy levels that are not required in most practical applications. For example, OLPSO and CPSO are able to give solutions that have an error range of about $1e-200$ or $1e-300$ \cite{Zhan11}, \cite{Bergh04}. Both these algorithms achieve this high degree of accuracy at the cost of significant computational overhead. We also note that in many applications, such high level of accuracy is generally not required as i) they operate in noisy environment and have a bound on accuracy due to the noise floor
%one can not hope to find an accurate solution below the noise floor
ii) a solution within a certain error margin %(e.g., $1e-7$ or $1e-5$)
would be considered sufficiently accurate and a lower error margin would result in minimal improvement in system performance.

Such a scenario frequently occurs in devices that have limited computational power (where finite machine precision would impose a limit on the level of accuracy that can be achieved) or limited power budget (for instance battery operated devices where it is desirable to minimize power consumption by performing only the necessary required computations). Similarly, many scenarios only require the estimate or solution to be accurate to a particular degree and a more accurate solution will not necessarily yield performance gain. For example, if the problem is to find the optimal weights of a neural network or optimal filter weights for an estimation problem or optimal distance calculations, then in engineering terms, a solution that has an accuracy of $1e{-5}$ might be as good as one that has an accuracy of $1e{-30}$.

Based on the analysis in the previous subsection, we compute the number of multiplications that would typically be needed by the standard PSO in a given scenario. For a typical swarm of $40$ particles optimizing a problem with say $30$ dimensions, a total of $5~\times~40~\times~30~=~ 6,000$ multiplications need to be performed per iteration for the update equations (\ref{x_orig}) and (\ref{v_inertia}). Even if the PSO algorithm runs for only $100$ iterations, then $6,000 \times 100 = 600,000$ multiplications need to be performed just for the update equations. Note that this number does not include the number of multiplications required for calculating the cost function for each particle at each iteration. %Our motivation is to design a strategy that would reduce the computational complexity of the update equations. This is of particular importance to applications that have a limited computational capacity or have a power constraint (e.g, cell phones, low cost wireless sensors etc.). The reduction in computations of the PSO algorithm permits these applications to utilize their resources more effectively elsewhere. To this end, we propose an event-triggering based approach described below in detail.

With this in view, we propose reducing the number of computations at the cost of acceptable performance degradation (we show in Section \ref{Results} that using our two proposed techniques, the performance degradation is negligible for all practical purposes). To this end, we propose an event-triggering-based approach described below in detail.

\subsection{The Event-Triggering Approach}

%For the $k$-th particle,
We rewrite (\ref{v_inertia}) as
\begin{equation}
v_{k}^{d} \left(i+1\right) =  w \cdot v_{k}^{d} \left(i\right) + \alpha(i) + \beta(i)
\label{v_inertia2}
\end{equation}
where $\alpha(i)$ and $\beta(i)$ represent the cognitive and social terms respectively. We note that the cognitive term consists of a constant $c_1$, random variable (drawn from a uniform distribution over the interval $[0,1]$) and the distance between the particle's current position and the local best position. This distance acts like a scaling factor that sets an upper limit on the maximum value of the cognitive term. The distance between the $d$-th dimension of the global best and the particle's current position has a similar role for the social term.

We define the particle's local distance for its $d$-th dimension as the absolute difference between the particle's current position and its local best position. The particle's global distance is defined in a similar fashion as the absolute difference between the particle's current position and its global best position, i.e.,
\begin{eqnarray}
{\rm local} \hspace{0.2cm} {\rm distance} &=& \left| p_k^d - x_k^d \right| \label{dist_local} \\
{\rm global} \hspace{0.2cm} {\rm distance} &=&\left| g^d - x_k^d \right| \label{dist_global}.
\end{eqnarray}
If the local distance is smaller than a certain preset threshold, denoted by $\gamma$, then the cognitive term in (\ref{v_inertia2}) is set to zero, i.e., it is not involved in the update process. The value of $\gamma$ is usually small. Mathematically

\begin{equation}
\label{loc_gamma}
\alpha(i) = \left\{\begin{array}{c c c}
c_1\cdot r_{1,k} (i) \cdot \left(p_{k}^{d} - x_{k}^{d} \left(i\right) \right) & & \left|p_{k}^{d} - x_{k}^{d} \left(i\right) \right| \geq \gamma\\ \\
 0  & & \left|p_{k}^{d} - x_{k}^{d} \left(i\right) \right| < \gamma
\end{array}\right.
\end{equation}

The idea here is that if the $d$-th dimension of the particle is very close to the $d$-th dimension of its local best, the contribution of the cognitive term of (\ref{v_inertia2}) will be very small, and hence can be ignored with negligible effect on the value of $v_{k}^{d} \left(i+1\right)$. The velocity of the particle for the $d$-th dimension would still be updated on the basis of its inertia weight term and the social term, $\beta$.

Similarly, if the $d$-th dimension of the particle is very close to the $d$-th dimension of the global best of the swarm then the contribution of the social term in the update of (\ref{v_inertia2}) would be negligible and can be ignored. That is %That is then the last term in (\ref{v_inertia}) is set to zero.
\begin{equation}
\label{glo_gamma}
\beta(i) = \left\{\begin{array}{c c c}
c_2\cdot r_{2,k}(i) \cdot \left(g^{d} - x_{k}^{d} \left(i\right) \right) & & \left|g^{d} - x_{k}^{d} \left(i\right) \right| \geq \gamma\\ \\
 0  & & \left|g^{d} - x_{k}^{d} \left(i\right) \right| < \gamma
\end{array}\right.
\end{equation}

%\begin{equation}
%\label{glo_gamma}
%\left|g^{d} - x_{k}^{d} \left(i\right) \right| < \gamma
%\end{equation}

Thus, the cognitive and social terms of the update equation for the $d$-th dimension will not be calculated if the particle's current position is within $\gamma$ distance of its local best position and the swarm's global best position, respectively. Note that these terms can be ignored in such a case owing to their negligible contributions to the update equation for this particular dimension. This can be viewed as an event-triggering approach. The particle's velocity in the $d$-th dimension is still updated due to the inertia weight term. Note that the proposed technique does not hinder the exploration by the particle in other dimensions. The technique merely avoids calculating those terms that would have negligible contribution to the velocity update (\ref{v_inertia}). Although we have considered the same value of $\gamma$ for all $D$ dimensions, different values can be used for each dimension if the scenario calls for different margin of errors for various parameters (dimensions) of the problem.

\section{Exploiting Separability and Dimension-wise Search}
\label{Dim-Wis}

In this section, we present the second proposed technique which aims to speed up the convergence.

\subsection{Motivation}

%We begin by explaining the motivation for this approach.
It is known that PSO suffers from the so called ``two steps forward, one step backward'' phenomenon \cite{Bergh04} as explained by the following example. Consider a $3$-dimensional sphere function with cost given as $f(x) = x_1^2 + x_2^2 + x_3^2$. Let the $k$-th particle's local best position be ${\textbf{p}}_k = [7, 5, -1]$ with cost $75$ and the position after the velocity update be ${\textbf{x}}_k = [6, 4, -3]$ with cost $61$. This would cause the $k$-th particle to update its local best position to ${\textbf{p}}_k = [6, 4, -3]$ even though the third dimension of the old local best vector was better than the third dimension of the new local best vector. This happens because the increased cost of the third dimension of ${\textbf{x}}_k$ is compensated by the gains of the first and the second dimensions; resulting in an overall lower cost of ${\textbf{x}}_k$. Although the overall cost has reduced, the third dimension is now further away from its optimal position. This motivates us to find an update rule that can avoid this ``two steps forward, one step backward'' phenomenon and somehow retain the ``good'' dimension during particle updates.

Another motivation for this approach is the simple fact that it is easier to deal with one dimension at a time; akin to the divide and conquer strategy. For problems with separable cost functions (discussed in the next subsection), each dimension can be optimized independently of the others. Although not true for every problem, it turns out that many important real-life problems have separable cost functions. This motivates us to investigate if separability can be exploited to aid the PSO algorithm for problems with separable cost functions.

Many time-critical applications require fast convergence. As such, the third motivation for this approach is to find a way to speed up the convergence of the PSO algorithm.

\subsection{Separable Cost Functions}

The cost functions in which each term depends on only a single dimension are termed as separable cost functions such as Sphere, Rastrigin and Sum-of-Powers functions. Thus, a $D$-dimensional Sphere function, given by $f(x) = x_1^2 + x_2^2 + \cdots + x_D^2$, can be separated into $D$ terms, with the $d$-th term being $x^2_d$, such that the summation of all $D$ terms gives the vector-cost of the sphere function. Some examples of applications with separable cost functions %that are also time-critical
include adaptive routing in mobile ad hoc networks \cite{DiCarro2005}, adaptive channel equalization \cite{Awami2007}-\cite{Yogi2010a}, multiuser detection \cite{Soo2007}, \cite{Choudhry2010}, adaptive beamforming \cite{Xiao2008}, \cite{Zaharis2011}, economic dispatch \cite{Gaing2003}-\cite{Chaturverdi2008}, biometrics \cite{Veera2005}, robotics \cite{Chatterjee2005}, and data clustering \cite{Yuwono2012,Merwe2003}. %Further applications can be found in \cite{Poli2007} and \cite{Poli2008}.

\subsection{Behavior of Standard PSO Algorithm}

In addition to the ``two steps forward, one step backward'' phenomenon, the characteristic behavior of the PSO algorithm also manifests itself in another way. Consider a standard PSO algorithm optimizing a $D$-dimensional problem. The update mechanism of the standard PSO updates the particle position vectors according to (\ref{x_orig}). The new particle position of the $k$-th particle, ${\textbf{x}}_k$, is compared to the particles local best, ${\textbf{p}}_k$. If the former has lower cost, ${\textbf{x}}_k$ becomes the new value of particle's local best while if the cost of the latter is less than that of the former, the current local best is retained. Similarly, the local best solution with the least cost among all particles is chosen as the global best solution.

These comparisons are based on the aggregate total cost calculations and are oblivious to the rich information contained by the particles on the single-dimension level. The cost associated with a particle might be lower for a solution p1 as compared to solution p2 and yet on the single-dimensional level, we may find that p2 has some values for certain dimensions that are more fit than their counter parts in solution p1. This happens because the normal update rule seeks to minimize the overall cost of the particle position vector and ignores the picture on the single-dimensional level. The idea is best explained by the following example.

We consider a $3$-dimensional sphere function with cost given as $f(x) = x_1^2 + x_2^2 + x_3^2$. Let the $k$-th particle's local best position be ${\textbf{p}}_k = [0, 7, 3]$, its position after the velocity update be ${\textbf{x}}_k = [8, 4, -1]$ and the global best position be ${\textbf{g}} = [-4, 2, 6]$. The corresponding costs are $58$, $81$ and $56$. As the updated position of the particle has a higher cost, $81$, than the particle's local best, $58$, the conventional total cost based update rule would retain the current local best. Also, the current local best vector of the $k$-th particle has a higher cost, $58$, than the global best vector, $56$, and thus, it will not effect the update of the global best vector. Thus, the particle will not learn anything new from ${\textbf{x}}_k$ in this generation.

However, if we observe carefully, both ${\textbf{p}}_k$ and ${\textbf{x}}_k$ contain useful information. The best vector that can be generated by combining components of ${\textbf{p}}_k$ and ${\textbf{x}}_k$ is $[0, 4, -1]$ which has a cost of $17$. The cost of this new vector is lower than the current particle best and hence it would become the new particle best. For simplicity, even if we neglect the particle best of all other particles and compare the global best only with the $k$-th particle's updated local best, we see that the optimum vector that can be formed by picking individual components of these solutions is $[0, 2, -1]$ which yields a minimum cost of $5$. Thus the standard PSO fails to use this rich information contained in individual dimensions.

\subsection{The Dimension-wise Approach}

One strategy to find the optimal local best can be to run a brute force exhaustive search to test all possible vectors that can be formed by combining the individual vector components of ${\textbf{p}}_k$ and ${\textbf{x}}_k$. Similarly, the strategy for the global best can be to run a similar exhaustive search over all local bests to find the global best vector. While this may be possible for the scenario when both the problem dimension, $D$, and the swarm size, $K$, are small, the strategy is utterly unfeasible for scenarios with large values of $D$ and $K$. It is worth mentioning here that the typical swarm size ($K=40$) is much too large to entertain the exhaustive search strategy. Interestingly, it turns out that if we focus on the class of problems that have separable cost functions, the optimum local/global best vector can be found at the same computational complexity as the standard PSO.

The dimension-wise cost evaluation strategy works as follows: Divide the swarm into $D$ $1$-dimensional swarms. Each swarm optimizes a single dimension. Run the regular PSO update equations (\ref{x_orig}) and (\ref{v_inertia}). Calculate the cost for the current position of the particle and store it in its component-value (dimension-wise) form. Use the component-value corresponding to each dimension to gauge its fitness. Based on this dimension-wise component-value, find the local and global best for each of the $D$ swarms. The $D$-dimensional vector obtained by the proposed method would be the same as the optimum vector obtained through exhaustive search of all possible combinations over $N$ particles. Mathematically, for a cost function
\begin{equation}
  f(x) = \sum_{n=0}^N g_n(x)
\end{equation}
we have
\begin{equation}
  \min f(x) = \sum_{n=0}^N \min g_n(x)
\end{equation}

We further explain the idea with the example of a $3$-dimensional sphere function, $f(x) = x_1^2 + x_2^2 + x_3^2$. As sphere is a separable cost function, the swarm is divided into three $1$-dimensional swarms. Let the $k$-th particle's local best position be ${\textbf{p}}_k = [0, 7, 3]$, its position after the velocity update be ${\textbf{x}}_k = [8, 4, -1]$. Comparing the component-cost terms for the first dimension, $0$ and $64$, we see that $p_{k,1}$ is the fitter dimension while comparing the component-cost terms for the second dimension, $49$ and $16$, we see that $x_{k,2}$ is the fitter dimension. Similarly for the third dimension, $x_{k,3}$ comes out to be the fitter dimension with a cost of $1$. The final local best vector, thus, comes out to be ${\textbf{p}}_{k} = [p_{k,1}, x_{k,2}, x_{k,3}]$, i.e., $[0,4,-1]$ with a cost of $17$. Again we assume the global best position to be ${\textbf{g}} = [-4, 2, 6]$. For simplicity, we neglect all the remaining $k-1$ particles and compare the component-cost terms of the current global best vector with the new local best of the $k$-th particle. We find that for the first dimension, the component-cost terms are $0$ and $16$ and thus $p_{k,1}$ comes out to be the fitter dimension. Similarly $g_2$ and $p_{k,3}$ are selected for the second and third dimension respectively. The new global best vector comes out to be ${\textbf{g}} = [p_{k,1}, g_2,  p_{k,3}] = [0,2,-1]$. Notice that in doing so, we did not incur any additional computational cost as the conventional method of calculating cost also calculates these component-cost terms and sums them together to find the final function-cost. It is also easy to see that the proposed approach not only avoids the ``two step forward, one step backward'' problem but also converges to the optimum solution much faster than PSO variants that use the conventional method for cost calculation. Thus, instead of performing an exhaustive search to find the optimum vector that can be generated from the given ${\textbf{p}}_k, {\textbf{x}}_k$ and ${\textbf{g}}$ vectors, the proposed approach allows us to find the same optimum vector at the same computational complexity as that of the standard PSO. A simple intelligent rearrangement of the intermediate calculations along with the separability constraint of the cost function allows us to greatly improve the performance of the PSO algorithm. Like the event-triggering approach, almost all of the existing PSO variants can benefit from this dimension-wise approach. Moreover, where almost all PSO variants suffer from the ``curse of dimensionality'', our proposed technique is very robust to the number of dimensions of the cost function. This is due to the divide and conquer nature of the proposed technique where each dimension is optimized individually.

It is important to point out the difference between the proposed approach and the approach in \cite{Bergh04}. At first glance, the two approaches might seem similar but both treat the problem in separate ways. The method of \cite{Bergh04} also divides the swarm of $N$ particles of $D$-dimensions into $D$ $1$-dimensional swarms, each with $N$ particles. Yet the cost is still calculated for a $D$-dimensional vector, known as the context vector. To test the fitness of a single dimension, the method calculates the cost function $N$ times. This requires a total of $DN$ cost function evaluations every iteration. Whereas our proposed method tests the cost function on a single dimension-wise basis by exploiting the dimension separability and performs only $N$ cost function evaluations per iteration. Although the approach in \cite{Bergh04} does have the advantage that it is not limited to problems with separable cost functions, our proposed method requires far less computations while achieving similar results in similar number of iterations. This means that although the particle's position is updated the same number of times, the function evaluations for our proposed method are much lesser than that of the method in \cite{Bergh04}.

\section{A Dimension-Wise Event-Triggering PSO Algorithm}
\label{combo}

\begin{algorithm}[b]
\textbf{The Dimension-Wise Event-Triggering PSO} \\
\emph{Step 1}. Divide the swarm into $D$ one-dimensional swarms. \\
\emph{Step 2}. Evaluate the local and global distances for each dimension of each particle using (\ref{dist_local}) and (\ref{dist_global}). \\
\emph{Step 3}. Update the dimensional velocity of each particle subject to
(\ref{loc_gamma}) and (\ref{glo_gamma}).\\
\emph{Step 4}. Evaluate the test function for each particle and store the
component-cost (corresponding to each dimension). \\
\emph{Step 5}. Update the local best value for each particle in each swarm
based on component-wise (dimensional) comparison. \\
\emph{Step 6}. Update the global best value for the swarm
based on component-wise (dimensional) comparison. \\
\emph{Step 8}. Stop if iteration limit is reached, otherwise go to step 2. \\
\caption{The proposed PSO-DE algorithm.}
\label{Combo}
\end{algorithm}

The previous two sections explained in detail the two proposed techniques. This section shows how the two proposed techniques can be combined to make the PSO algorithm computationally less complex as well as converge faster. We term the modified PSO algorithm as a dimension-wise event-triggered PSO (PSO-DE) algorithm.

The event-triggering approach applied threshold values on the two distances given by (\ref{dist_local}) and (\ref{dist_global}). The dimension-wise approach transformed the $D$-dimensional hyperspace into a collection of $D$ $1$-dimensional spaces, which are searched individually. This results in a vastly improved performance at a low computational cost as shown in Section \ref{Results}. Algorithm \ref{Combo} outlines the proposed PSO-DE algorithm.

\section{Results and Discussion}
\label{Results}

In this section, the proposed techniques are applied on some PSO variants to substantiate our claims. The algorithms are tested on five different test functions. Two different hyperspaces are used, with $30$ and $60$ dimensions respectively. A swarm of $40$ particles is used. Parameter values of the PSO variants tested are taken from their respective references. %\emph{\textbf{In order to have a fair comparison, each PSO variant is run for $5000$ iterations, which results in the same number of function evaluations for all }} variants considered in this work.
Each PSO variant is run for $5000$ iterations and the results are averaged over $50$ experiments. The threshold parameter, $\gamma$, is set to $1e-7$ for the proposed event-triggering approach. Several performance measures are compared in order to judge the performance of the proposed algorithms against other PSO variants. Specifically, we compare the mean of the test function returned by the algorithm, number of iterations needed for convergence, computational complexity and success rate.
%we compare the mean values of the test functions returned by the algorithms. We also compare the time required by each algorithm to reach its final value.
We consider an algorithm to have successfully converged for a test function if the final value returned by the algorithm is less than a user specified threshold called the ``accept value'' (the accept value for each test function is given in Table \ref{TestTable}). We define the success rate (SR) of an algorithm as the percentage of runs that converged successfully.

\begin{table*}[t]
\begin{center}
\begin{tabular}{c l l l l l l} \hline
 & {\bf Function names} & {\bf Function equations} & {\bf Search range} & {\bf Initialization range} & \multicolumn{2}{c}{{\bf Accept}}\\
 & & & & & D=30& D=60\\
\hline
1. & {\em Sphere} & $f_1 (x) = \sum\limits_{i = 1}^D {x_i^2}$ & $[-100,100]^D$ & $[-100,50]^D$ & $1$ & $1$\\
2. & {\em Rosenbrock} & $f_2 (x) = \sum\limits_{i = 1}^D {\left( {100{{\left( {x_i^2 - {x_{i + 1}}} \right)}^2} + {{\left( {x_i - 1} \right)}^2}} \right)}$ & $[-10, 10]^D$ & $[-10, 10]^D$ & $200$ & $500$\\
3. & {\em Rastrigin} & $f_3 (x) = \sum\limits_{i = 1}^D {\left(x_i^2 - 10{\rm cos}\left(2\pi x_i\right) + 10\right)}$ & $[-5.12,5.12]^D$ & $[-5.12,2]^D$ & $100$& $200$\\
4. & {\em Michalewicz} & $f_4 (x) = - \sum\limits_{i = 1}^D {\rm sin} x_i \left( {\rm sin} \frac{ix_i^2}{\pi} \right)^{2m}$, $m=10$ & $[-10, 10]^D$ & $[-10, 10]^D$ & $1$ &  $1$\\
5. & {\em Sum-of-Powers} & $f_5 (x) = \sum\limits_{i = 1}^D {\left|x\right|^{i+1}}$ & $[-10, 10]^D$ & $[-10, 10]^D$ & $1$ & $1$\\
\hline
\end{tabular}
\end{center}
%\vspace{-0.2cm}
\caption{Test Functions}
\label{TestTable}
\end{table*}

\begin{table*}[t]
\begin{center}
\begin{tabular}{l c c c c  c c c c}
\hline
& \multicolumn{4}{|c|}{\textbf{Results for 30 Dimensions}} & \multicolumn{4}{c}{\textbf{Results for 60 Dimensions}}\\
\hspace{0.3cm} \textbf{f(x)} & \multicolumn{1}{|c}{\textbf{PSO}} & \textbf{PSO-D} & \textbf{PSO-E} & \multicolumn{1}{c|}{\textbf{PSO-DE}}  & \textbf{PSO} & \textbf{PSO-D} & \textbf{PSO-E} & \textbf{PSO-DE}\\
\hline  \\ [-1.5ex]
\hspace{0.3cm} Mean & $1.25e-32$ & $5.08e-308$ & $7.18e-14$ & $7.86e-21$ & $1.15e-10$ & $1.07e-307$ & $1.52e-08$ & $1.24e-20$\\
\hspace{0.3cm} Iters. & $5000$ & $4717$ & $4069$ & $1771$ & $5000$ & $4725$ & $5000$ & $1889$\\
$f_1$ Comp. & $100\%$ & $100\%$ & $79.37\%$ & $43.46\%$ & $100\%$ & $100\%$ & $96.10\%$ & $43.47\%$ \\
\hspace{0.3cm} SR & $100\%$ & $100\%$ & $100\%$ & $100\%$ & $100\%$ & $100\%$ & $100\%$ & $100$\\
[1ex]  \\ [-1.5ex]
\hspace{0.3cm} Mean & $25.2$ & $0$ & $29.3$ & $5.07e-13$ & $6.06$ & $1.67e-308$ & $3.57$ & $3.81e-23$\\
\hspace{0.3cm} Iters.& $5000$ & $3217$ & $5000$ & $2545$ & $3598$ & $3369$ & $5000$ & $1410$\\
$f_2$ Comp. & $100\%$ & $100\%$ & $97.76\%$ & $51.47\%$ & $100\%$ & $100\%$ & $97.39\%$ & $53.64\%$\\
\hspace{0.3cm} SR & $98\%$ & $100\%$ & $100\%$ & $100\%$ & $66\%$ & $100\%$ & $56\%$ & $100\%$\\
 [1ex]  \\ [-1.5ex]
\hspace{0.3cm} Mean & $47.75$ & $0$ & $44.6$ & $0$ & $133$ & $0$ & $131$ & $0$\\
\hspace{0.3cm} Iters. & $4848$ & $1114$ & $4289$ & $1197$ & $5000$ & $1078$ & $5000$ & $1183$\\
$f_3$ Comp. & $100\%$ & $100\%$ & $78.22\%$ & $42.26\%$ & $100\%$ & $100\%$ & $93.15\%$ & $42.27\%$\\
\hspace{0.3cm} SR & $100\%$ & $100\%$ & $100\%$ & $100\%$ & $100\%$ & $100\%$ & $98\%$ & $100\%$\\
 [1ex]  \\ [-1.5ex]
\hspace{0.3cm} Mean & $1.33e-14$ & $2.65e-254$ & $2.41e-07$ & $2.89e-126$ & $1.06e-07$ & $3.89e-45$ & $9.14e-07$ & $1.30e-45$\\
\hspace{0.3cm} Iters. &  $2335$ & $2337$ & $2268$ & $3198$ & $2720$ & $248$ & $2335$ & $333$\\
$f_4$ Comp. & $100\%$ & $100\%$ & $97.78\%$ & $44.89\%$ & $100\%$ & $100\%$ & $97.92\%$ & $48.51\%$\\
\hspace{0.3cm} SR & $100\%$ & $100\%$ & $100\%$ & $100\%$ & $100\%$ & $100\%$ & $100\%$ & $100\%$\\
 [1ex]  \\ [-1.5ex]
\hspace{0.3cm} Mean & $2.19e-51$ & $1.36e-308$ & $3.34e-17$ & $1.65e-22$ & $3.51e-15$ & $1.60e-308$ & $1.06e-15$ & $7.85e-23$\\
\hspace{0.3cm} Iters. & $5000$ & $4541$ & $3910$ & $1194$ & $5000$ & $4175$ & $5000$ & $1438$\\
$f_5$ Comp. & $100\%$ & $100\%$ & $90.51\%$ & $42.56\%$ & $100\%$ & $100\%$ & $97.09\%$ & $53.63\%$\\
\hspace{0.3cm} SR & $90\%$ & $100\%$ & $82\%$ & $100\%$ & $46\%$ & $100\%$ & $42\%$ & $100\%$\\
\hline
\end{tabular}
\end{center}
%\vspace{-0.2cm}
\caption{Performance comparison of proposed techniques applied to standard PSO algorithm.}
\label{PSO2DE1}
\end{table*}

Specifically, we applied the proposed techniques to the standard PSO, DMSPSO \cite{Liang05}, CLPSO \cite{Liang06} and HPSO \cite{Ratnaweera04} algorithms and compared the performance of these modified variants with the original algorithms. These algorithms are chosen as an example and the proposed techniques can be applied to other PSO algorithms in literature.

%To begin with, the mean values for each of the test functions are compared and the results are discussed. Next, we present a study of reduction in computation complexity as well as the number of iterations when the proposed techniques are applied to the PSO algorithm. Since different PSO variants behave differently, we also provide a detailed study of their performance comparison them not only with each other but also individually (after applying the proposed technique) in order to see how the proposed techniques can aid to improve the convergence speed of these variants as well as reduce computations.

\subsection{Test Functions}

The test functions used in our  experiments are given in Table \ref{TestTable}. We use test functions of varying complexity in the experiments in order to test the robustness of the proposed techniques. An important point to note is that all test functions chosen here are either completely separable or ``almost'' separable. %Such functions are deliberately chosen here because several time-critical PSO applications have similar cost functions.
The focus here is on separable functions as many important problems have cost functions that are separable (as detailed in Section \ref{Dim-Wis}).
%\emph{\textbf{, e.g., adaptive beamforming, adaptive routing in mobile ad hoc networks, system identification, adaptive channel equalization, multiuser detection, robotics, economic dispatch and biometrics.}}

Function $f_1$ is unimodal while $f_2, f_3, f_4$ and $ f_5$ are multimodal. Note that the Rosenbrock function can be treated as a multimodal function in high-dimensions \cite{Shang2006}. Table \ref{TestTable} also gives the accept values for these functions. If the solution returned by an algorithm is higher than the accept value, the algorithm is considered to have failed to converge.

\begin{table*}[t]
\begin{center}
\begin{tabular}{ l c c c c  c c c c}
\hline
& \multicolumn{4}{|c|}{\textbf{Results for 30 Dimensions}} & \multicolumn{4}{c}{\textbf{Results for 60 Dimensions}}\\
\hspace{0.3cm} \textbf{f(x)} & \multicolumn{1}{|c}{\textbf{DMS}} & \textbf{DMS-D} & \textbf{DMS-E} & \multicolumn{1}{c|}{\textbf{DMS-DE}} & \textbf{DMS} & \textbf{DMS-D} & \textbf{DMS-E} & \textbf{DMS-DE}\\
\hline  \\ [-1.5ex]
\hspace{0.3cm} Mean & $4.73e-17$ & $5.11e-308$ & $9.57e-14$ & $5.26e-21$ & $6.72e-05$ & $1.03e-307$ & $1.46e-03$ & $1.15e-20$\\
$f_1$ Iters. &  $5000$ & $4215$ & $4969$ & $2004$ & $5000$ & $4233$ & $5000$ & $2142\%$\\
\hspace{0.3cm} Comp. &  $100\%$ & $100\%$ & $90.80\%$ & $43.48\%$ & $100\%$ & $100\%$ & $95.38\%$ & $43.45\%$\\
\hspace{0.3cm} SR & $100\%$ & $100\%$ & $100\%$ & $100\%$ & $100\%$ & $100\%$ & $100\%$ & $100\%$\\
[1ex]  \\ [-1.5ex]
\hspace{0.3cm} Mean & $24.2$ & $0$ & $31.9$ & $2.55e-13$ & $64.6$ & $0$ & $97.6$ & $5.41e-13$\\
$f_2$ Iters. & $5000$ & $4147$ & $5000$ & $3443$ & $5000$ & $4093$ & $5000$ & $3266$\\
\hspace{0.3cm} Comp. & $100\%$ & $100\%$ & $95.43\%$ & $54.58\%$ & $100\%$ & $100\%$ & $95.63\%$ & $53.81\%$\\
\hspace{0.3cm} SR & $100\%$ & $100\%$ & $100\%$ & $100\%$ & $100\%$ & $100\%$ & $90\%$ & $100\%$\\
 [1ex]  \\ [-1.5ex]
\hspace{0.3cm} Mean & $16.8$ & $0$ & $27.8$ & $0$ & $50.4$ & $0$ & $92$ & $0$\\
$f_3$ Iters. & $5000$ & $1172$ & $5000$ & $1172$ & $5000$ & $1139$ & $5000$ & $1185$\\
\hspace{0.3cm} Comp. & $100\%$ & $100\%$ & $96.67\%$ & $42.29\%$ & $100\%$ & $100\%$ & $96.78\%$ & $42.28\%$\\
\hspace{0.3cm} SR & $100\%$ & $100\%$ & $100\%$ & $100\%$ & $100\%$ & $100\%$ & $100\%$ & $100\%$\\
 [1ex]  \\ [-1.5ex]
\hspace{0.3cm} Mean & $1.61e-15$ & $4.09e-309$ & $4.28e-08$ & $4.72e-309$ & $6.72e-15$ & $7.66e-309$ & $1.23e-07$ & $6.51e-156$\\
$f_4$ Iters. & $4437$ & $1111$ & $4476$ & $1135$ & $4458$ & $1445$ & $4504$ & $1116$\\
\hspace{0.3cm} Comp. & $100\%$ & $100\%$ & $97.29\%$ & $43.37\%$ & $100\%$ & $100\%$ & $97.19\%$ & $44.22\%$\\
\hspace{0.3cm} SR & $100\%$ & $100\%$ & $100\%$ & $100\%$ & $100\%$ & $100\%$ & $100\%$ & $100\%$\\
 [1ex]  \\ [-1.5ex]
\hspace{0.3cm} Mean & $7.51e-28$ & $1.18e-308$ & $5.99e-17$ & $1.79e-22$ & $1.07e-07$ & $1.61e-308$ & $1.33e-01$ & $4.53e-22$\\
$f_5$ Iters. & $5000$ & $4144$ & $4925$ & $1806$ & $5000$ & $5000$ & $5000$ & $1569$\\
\hspace{0.3cm} Comp. & $100\%$ & $100\%$ & $93.95\%$ & $42.54\%$ & $100\%$ & $100\%$ & $95.44\%$ & $56.02\%$\\
\hspace{0.3cm} SR & $100\%$ & $100\%$ & $100\%$ & $100\%$ & $100\%$ & $100\%$ & $78\%$ & $100\%$\\
\hline
\end{tabular}
\end{center}
%\vspace{-0.2cm}
\caption{Performance comparison of proposed techniques applied to DMSPSO algorithm.}
\label{DMSDE1}
\end{table*}

\subsection{Modified PSO Algorithm}

Table \ref{PSO2DE1} shows the results obtained when the proposed techniques are applied to the standard PSO algorithm. In the table, PSO stands for the standard PSO algorithm, PSO-D is the combination of the standard PSO algorithm and the dimension-wise technique, PSO-E is the combination of standard PSO with the event-triggering approach while PSO-DE is the combination of both the proposed techniques with the standard PSO algorithm. The row labeled ``Mean'' shows the mean value of the function evaluated over successful runs (a run is deemed successful if the algorithm achieves the ``accept'' value for the function given in Table \ref{TestTable}). The label ``Iters.'' stands for the number of iterations required by the algorithm to converge to its mean value. The label ``Comp.'' shows the computational complexity. It is measured in terms of the computational complexity of the base algorithm to which the proposed techniques are applied. For Table \ref{PSO2DE1}, the base algorithm is the standard PSO algorithm. Thus, if Table \ref{PSO2DE1} has an entry for an algorithm with a computational complexity of $100\%$, it means the algorithm in question performs the same number of computations as the base algorithm. Similarly, a value of $60\%$ means the number of calculation performed by this algorithm is $60\%$ of that performed by the base algorithm. The label ``SR'' gives the success rate, i.e., the percentage of runs for which the algorithm achieves the ``accept'' value. The scenario simulated here is the following: Given all the algorithms are to run for a fixed duration of time (time constraint), what is the performance of each algorithm with respect to power consumption (measured by computational complexity), accuracy (measured by mean value) and convergence (measured by success rate)?

Considering the results for the $30$-D case in Table \ref{PSO2DE1}, we see that the PSO-D variant significantly enhances the mean performance of the PSO algorithm. It also speeds up the convergence of the algorithm. The PSO-E variant has slightly lower computational complexity as compared to the standard PSO at the cost of degraded mean performance. The results of the PSO-DE variant are of particular interest. The results show that with the exception of $f_4$, PSO-DE not only converges faster than the PSO algorithm for all test functions but also has a lower mean value along with a significant reduction in computational cost. For function $f_4$, PSO-DE has a much lower cost ($2.89e-126$ as opposed to $1.33e-14$) and lower computational complexity ($44.89\%$), although it takes a bit longer than standard PSO ($3198$ iterations as opposed to $2335$) to reach its final value. Nevertheless, it still requires less computations than the PSO algorithm. The result for $f_2$ and $f_3$ are also significant in that both the PSO-D and PSO-DE variants significantly improve the mean value performance of the original PSO algorithm. Similar trends are observed for the $60$-D case. We also note that both the speed of convergence and the reduction in computation of the PSO-DE algorithm exhibit robustness to increase in the number of dimensions.

\subsection{A Modified DMSPSO Algorithm}

Table \ref{DMSDE1} shows the $30$-D and $60$-D results when the proposed techniques are applied to the DMSPSO algorithm \cite{Liang05}. In this table, the computational complexity is calculated with respect to the original DMSPSO algorithm. We observe similar trends as in the case of the standard PSO algorithm. The DMS-D algorithm has a much improved mean performance as compared to the DMSPSO algorithm as well as much faster convergence. The DMS-E algorithm offers some reduction in the computational complexity (approximately $3\%$ to $9\%$) at the cost of degraded performance. The DMS-DE algorithm, however, offer much faster convergence ($2$ to $4$ times faster) and much lower mean performance (except for $f_5$) at only about half the computational cost. The mean performance of the PSO-DE algorithm is particularly significant for $f_2$ and $f_3$ as these two functions are considered very hard to optimize. For the DE variant, the mean performance of the DMS algorithm is improved from $24.2$ and $16.8$ to $2.55e-13$ and $0$ for these two functions, respectively.

\begin{table*}[t]
\begin{center}
\begin{tabular}{ l c c c c  c c c c}
\hline
& \multicolumn{4}{|c|}{\textbf{Results for 30 Dimensions}} & \multicolumn{4}{c}{\textbf{Results for 60 Dimensions}}\\
\hspace{0.3cm} \textbf{f(x)} & \multicolumn{1}{|c}{\textbf{CLPSO}} & \textbf{CLPSO-D} & \textbf{CLPSO-E} & \multicolumn{1}{c|}{\textbf{CLPSO-DE}} & \textbf{CLPSO} & \textbf{CLPSO-D} & \textbf{CLPSO-E} & \textbf{CLPSO-DE}\\
\hline \\ [-1.5ex]
\hspace{0.3cm} Mean & $1.67e-309$ & $5.52e-308$ & $6.03e-14$ & $6.48e-20$ & $2.32e-309$ & $1.07e-307$ & $1.39e-13$ & $1.24e-19$\\
$f_1$ Iters. &  $4047$ & $4012$ & $1210$ & $1073$ & $4088$ & $3964$ & $1150$ & $1011$\\
\hspace{0.3cm} Comp. & $100\%$ & $100\%$ & $42.48\%$ & $42.40\%$ & $100\%$ & $100\%$ & $42.45\%$ & $ 42.37\%$\\
\hspace{0.3cm} SR & $100\%$ & $100\%$ & $100\%$ & $100\%$ & $100\%$ & $100\%$ & $100\%$ & $100\%$\\
 [1ex]  \\ [-1.5ex]
\hspace{0.3cm} Mean & $28.4$ & $163$ & $28.5$ & $162$ & $58.1$ & $ 372 $ & $58.1$ & $397$\\
$f_2$ Iters. & $653$ & $407$ & $634$ & $418$ & $4569$ & $4744$ & $1615$ & $4520$\\
\hspace{0.3cm} Comp. & $100\%$ & $100\%$ & $43.59\%$ & $43.58\%$ & $100\%$ & $100\%$ & $43.11\%$ & $43.10\%$\\
\hspace{0.3cm} SR & $100\%$ & $68\%$ & $100\%$ & $56\%$ & $100\%$ & $84\%$ & $100\%$ & $88\%$\\
 [1ex]  \\ [-1.5ex]
\hspace{0.3cm} Mean & $0$ & $0$ & $1.16e-11$ & $0$ & $5.78e-14$ & $0$ & $1.16e-11$ & $0$\\
$f_3$ Iters. & $935$ & $840$ & $1048$ & $747$ & $1014$ & $858$ & $1105$ & $756$\\
\hspace{0.3cm} Comp. & $100\%$ & $100\%$ & $41.63\%$ & $41.53\%$ & $100\%$ & $100\%$ & $41.58\%$ & $41.50\%$\\
\hspace{0.3cm} SR & $100\%$ & $100\%$ & $100\%$ & $100\%$ & $100\%$ & $100\%$ & $100\%$ & $100\%$\\
 [1ex]  \\ [-1.5ex]
\hspace{0.3cm} Mean & $4.27e-295$ & $4.57e-309$ & $2.07e-205$ & $4.12e-309$ & $2.41e-11$ & $7.46e-309$ & $1.11e-12$ & $2.26e-204$\\
$f_4$ Iters. & $2275$ & $985$ & $2128$ & $999$ & $630$ & $1017$ & $480$ & $1010$\\
\hspace{0.3cm} Comp. & $100\%$ & $100\%$ & $43.17\%$ & $41.77\%$ & $100\%$ & $100\%$ & $42.61\%$ & $41.73\%$\\
\hspace{0.3cm} SR & $100\%$ & $100\%$ & $100\%$ & $100\%$ & $100\%$ & $100\%$ & $100\%$ & $100\%$\\
 [1ex]  \\ [-1.5ex]
\hspace{0.3cm} Mean & $1.71e-309$ & $1.50e-308$ & $2.23e-21$ & $2.05e-21$ & $1.66e-309$ & $1.77e-308$ & $4.85e-21$ & $1.26e-21$\\
$f_5$ Iters. & $4017$ & $3991$ & $1119$ & $998$ & $3965$ & $3956$ & $1231$ & $1119$\\
\hspace{0.3cm} Comp. & $100\%$ & $100\%$ & $41.77\%$ & $41.74\%$ & $100\%$ & $100\%$ & $41.72\%$ & $41.71\%$\\
\hspace{0.3cm} SR & $100\%$ & $100\%$ & $100\%$ & $100\%$ & $94\%$ & $100\%$ & $88\%$ & $100\%$\\
\hline
\end{tabular}
\end{center}
%\vspace{-0.2cm}
\caption{Performance comparison of proposed techniques applied to CLPSO algorithm.}
\label{CLDE1}
\end{table*}

\begin{table*}[t]{\scriptsize
\begin{center}
\begin{tabular}{l c c c c c c c c c c}
\hline
& \multicolumn{5}{|c|}{\textbf{Results for 30 Dimensions}} & \multicolumn{5}{c}{\textbf{Results for 60 Dimensions}}\\
\hspace{0.3cm} \textbf{f(x)} & \multicolumn{1}{|c}{\textbf{HP}} & \textbf{HP-D} & \textbf{HP-E} & \textbf{HP-DE} & \textbf{HP-DE2} & \multicolumn{1}{|c}{\textbf{HP}} & \textbf{HP-D} & \textbf{HP-E} & \textbf{HP-DE} & \textbf{HP-DE2}\\
\hline
\hspace{0.3cm} Mean & $1.44e\hspace{-.2em}-\hspace{-.2em}36$ & $6.04e\hspace{-.2em}-\hspace{-.2em}308$ & $7.69e\hspace{-.2em}-\hspace{-.2em}11$ & $8.71e\hspace{-.2em}-\hspace{-.2em}21$ & $1.33e\hspace{-.2em}-\hspace{-.2em}14$ & $3.99e\hspace{-.2em}-\hspace{-.2em}22$ & $1.17e\hspace{-.2em}-\hspace{-.2em}307$ & $8.48e\hspace{-.2em}-\hspace{-.2em}09$ & $1.59e\hspace{-.2em}-\hspace{-.2em}20$ & $2.64e\hspace{-.2em}-\hspace{-.2em}14$\\
$f_1$ Iters. & $5000$ & $3425$ & $4987$ & $4997$ & $85$ & $5000$ & $3594$ & $5000$ & $4999$ & $86$ \\
\hspace{0.3cm} Comp. & $100\%$ & $100\%$ & $85.16\%$ & $95.36\%$ & $0.62\%$ & $100\%$ & $100\%$ & $86\%$ & $95.36\%$ & $0.62\%$\\
\hspace{0.3cm} SR & $100\%$ & $100\%$ & $100\%$ & $100\%$ & $100\%$ & $100\%$ & $100\%$ & $100\%$ & $100\%$ & $100\%$\\
\hline
\hspace{0.3cm} Mean & $25.5$ & $6.93e\hspace{-.2em}-\hspace{-.2em}11$ & $23.8$ & $1.12e\hspace{-.2em}-\hspace{-.2em}12$ & $9.62e\hspace{-.2em}-\hspace{-.2em}06$ & $73.8$ & $3.24e\hspace{-.2em}-\hspace{-.2em}10$ & $81.5$ & $2.54e\hspace{-.2em}-\hspace{-.2em}12$ & $3.65e\hspace{-.2em}-\hspace{-.2em}2$\\
$f_2$ Iters. & $5000$ & $4244$ & $5000$ & $4995$ & $587$ & $5000$ & $3791$ & $5000$ & $5000$ & $1956$\\
\hspace{0.3cm} Comp. & $100\%$ & $100\%$ & $78.32\%$ & $93.94\%$ & $14.35\%$ & $100\%$ & $100\%$ & $80.94\%$ & $94\%$ & $14.35\%$\\
\hspace{0.3cm} SR & $100\%$ & $100\%$ & $100\%$ & $100\%$ & $100\%$ & $100\%$ & $100\%$ & $100\%$ & $100\%$ & $100\%$\\
\hline
\hspace{0.3cm} Mean & $15.3$ & $0$ & $5.06$ & $0$ & $2.06e\hspace{-.2em}-\hspace{-.2em}12$ & $42.3$ & $0$ & $22.3$ & $0$ & $4.05e\hspace{-.2em}-\hspace{-.2em}12$\\
$f_3$ Iters. & $5000$ & $151$ & $5000$ & $230$ & $1002$ & $5000$ & $153$ & $5000$ & $206$ & $898$ \\
\hspace{0.3cm} Comp. & $100\%$ & $100\%$ & $92.84\%$ & $94.93\%$ & $3.12\%$ & $100\%$ & $100\%$ & $92.17\%$ & $94.93\%$ & $3.11\%$\\
\hspace{0.3cm} SR & $100\%$ & $100\%$ & $100\%$ & $100\%$ & $100\%$ & $100\%$ & $100\%$ & $100\%$ & $100\%$ & $100\%$\\
\hline
\hspace{0.3cm} Mean & $3.20e\hspace{-.2em}-\hspace{-.2em}06$ & $3.49e\hspace{-.2em}-\hspace{-.2em}264$ & $1.81e\hspace{-.2em}-\hspace{-.2em}05$ & $4.14e\hspace{-.2em}-\hspace{-.2em}153$ & $4.99e\hspace{-.2em}-\hspace{-.2em}87$ & $1.59e\hspace{-.2em}-\hspace{-.2em}05$ & $6.45e\hspace{-.2em}-\hspace{-.2em}251$ & $2.12e\hspace{-.2em}-\hspace{-.2em}05$ & $2.60e\hspace{-.2em}-\hspace{-.2em}139$ & $1.06e\hspace{-.2em}-\hspace{-.2em}88$\\
$f_4$ Iters. & $4996$ & $1194$ & $4964$ & $4571$ & $3276$ & $4965$ & $1447$ & $4918$ & $4562$ & $2709$ \\
\hspace{0.3cm} Comp. & $100\%$ & $100\%$ & $99.72\%$ & $95.73\%$ & $19.01\%$ & $100\%$ & $100\%$ & $99\%$ & $96.07\%$ & $22.74\%$\\
\hspace{0.3cm} SR & $100\%$ & $100\%$ & $100\%$ & $100\%$ & $100\%$ & $100\%$ & $100\%$ & $100\%$ & $100\%$ & $100\%$\\
\hline
\hspace{0.3cm} Mean & $1.61e\hspace{-.2em}-\hspace{-.2em}29$ & $5.46e\hspace{-.2em}-\hspace{-.2em}308$ & $8.08e\hspace{-.2em}-\hspace{-.2em}16$ & $1.72e\hspace{-.2em}-\hspace{-.2em}22$ & $4.31e\hspace{-.2em}-\hspace{-.2em}16$ & $4.34e\hspace{-.2em}-\hspace{-.2em}25$ & $7.86e\hspace{-.2em}-\hspace{-.2em}308$ & $1.33e\hspace{-.2em}-\hspace{-.2em}15$ & $3.83e\hspace{-.2em}-\hspace{-.2em}22$ & $4.48e\hspace{-.2em}-\hspace{-.2em}16$\\
$f_5$ Iters. & $5000$ & $3493$ & $5000$ & $4991$ & $66$ & $5000$ & $3182$ & $4998$ & $4886$ & $65$\\
\hspace{0.3cm} Comp. & $100\%$ & $100\%$ & $91.20\%$ & $94.89\%$ & $0.55\%$ & $100\%$ & $100\%$ & $87.49\%$ & $93.04\%$ & $17.89\%$\\
\hspace{0.3cm} SR & $100\%$ & $100\%$ & $100\%$ & $100\%$ & $100\%$ & $100\%$ & $100\%$ & $100\%$ & $100\%$ & $100\%$\\
\hline
\end{tabular}
\end{center}
%\vspace{-0.2cm}
\caption{Performance comparison of proposed techniques applied to HPSO algorithm.}
\label{HPDE1}}
\end{table*}

In the $60$-D case, the DMS-E algorithm has a SR of $90\%$ and $78\%$ for Rosenbrock ($f_2$) and Sum of Powers ($f_5$) functions respectively. The reason behind this is that the event-triggering approach, in order to save computations, skips calculating the cognitive and social terms if the particle's current position is close to its local or global best respectively. As a result, its mean performance will be less accurate than the base algorithm. The DMS algorithm itself has a mean error performance of $64.4$ for the Rosenbrock function so it is expected that the DMS-E will have lower mean performance than the DMSPSO algorithm. Similarly, the Sum of Powers function is known to be difficult to optimize. High dimensionality coupled with the highly complex nature of these functions are the cause of less than $100\%$ SR of the DMS-E algorithm.

The DMS-DE variant, however, performs very well for all functions including the Rosenbrock and Sum of Powers. It gives a much better mean performance than the DMS algorithm for almost all test cases; with less than half the computations and significant reduction ($2 - 4$ times) in convergence time.

\subsection{A Modified CLPSO Algorithm}

Table \ref{CLDE1} shows the results obtained by applying the two techniques to the CLPSO algorithm \cite{Liang06}. We note that the CLPSO has been shown to have remarkable performance. The CLPSO-E variant offers saving in computational complexity (the computational complexity is measured relative to the original CLPSO algorithm) at the cost of lower mean performance. The performance of the CLPSO-DE algorithm is similar with its key features being faster convergence and lower computational cost at the sacrifice of superior mean performance.  The mean performance of the CLPSO-D variant is comparable to the CLPSO algorithm. This is in contrast to the other algorithms tested (standard PSO, DMSPSO, and HPSO) where the D variant of those algorithms considerably improved their mean performance. The reason behind this is the highly randomized search behavior of the CLPSO where each particle learns from different particles for its different dimensions resulting in superior performance and compared to other PSO algorithms.

%\textbf{Neither CLPSO nor its proposed variants perform well on the Rosenbrock function. In fact, the performance of the D and DE variants is much poorer than the original CLPSO algorithm for the Rosenbrock function. It is know that CLPSO does not perform well for the Rosenbrock function \cite{Liang06} so the performance of CLPSO-E is also poor given that the event-triggering variant tends have relatively higher mean error than its original algorithm.} %Also, the Rosenbrock function is not completely separable (each individual term of the cost function depends on $2$ dimensions). This, coupled with the poor performance of the }

\begin{table*}[t]
\begin{center}
\begin{tabular}{ l l c c c c c c c c c}
\hline \\ [-1.75ex]
&\textbf{f(x)} & \textbf{Threshold} & \textbf{PSO} & \textbf{DMS} & \textbf{CLPSO} & \textbf{HPSO} & \textbf{PSO-DE} & \textbf{DMS-DE} & \textbf{CLPSO-DE} & \textbf{HPSO-DE2}\\
\hline \\ [-1ex]
&\multirow{2}{*}{$f_1$}&$1e-10$ & $3685$ & $4633$ & $762$ & $2025$ & $ 483$ & $ 424$ & $ 389 $ & $44$\\
&                      &$1e-15$ & $4044$ & $4884$ & $913$ & $2600$ & $ 768$ & $ 675$ & $ 665 $  & $\times$\\ [1ex]  \\ [-1.5ex]
&\multirow{2}{*}{$f_2$}&$1e-10$ & $\times$ & $ \times$ & $ \times$ & $\times$ &  $ 1806$ & $ 2115$ & $ \times $ & $\times$\\
&               &$1e-15$ & $\times$ & $ \times$ & $ \times $ & $\times$ &  $\times$ & $ \times$ & $ \times $ & $ \times $\\ [1ex]  \\ [-1.5ex]
&\multirow{2}{*}{$f_3$}&$1e-10$ & $\times $ & $\times $ & $763$ & $\times$ &  $ 463$ & $ 663$ & $ 385 $ & $83$\\
&                      &$1e-15$ & $\times$ & $ \times$ & $ 870$ & $\times$ &  $ 822$ & $ 1055$ & $ 590 $ & $\times$\\ [1ex]  \\ [-1.5ex]
\raisebox{4ex}[1.5pt] {$\begin{array}{c}
\mbox{Results}\\ \mbox{for} \\ \mbox{$30$-\emph{D}}
\end{array}$}&\multirow{2}{*}{$f_4$}&$1e-10$ & $1923$ & $ 1906$ & $ 224 $ & $\times$ & $2$ & $ 2$ & $ 2 $ & $2$\\
&                      &$1e-15$ & $2072(\textbf{40\%})$ & $ 2432 (\textbf{74\%})$ & $ 260$ & $\times$ &  $ 3$ & $ 2$ & $ 3 $ & $3$\\ [1ex]  \\ [-1.5ex]
&\multirow{2}{*}{$f_5$}&$1e-10$ & $2887(\textbf{90\%})$ & $ 3777$ & $ 514$ & $1086$ &  $ 20$ & $ 31$ & $ 245 $ & $32$\\
&                      &$1e-15$ & $3278(\textbf{90\%})$ & $ 4400$ & $ 720$ & $2223$ &  $ 65$ & $ 83$ & $ 559 $ & $50$\\ [.5ex] \hline \\ [-1.5ex]
&\multirow{2}{*}{$f_1$}&$1e-10$ & $4931(\textbf{68\%})$ & $\times$ & $805$ & $3037$ & $582$ & $550$ & $384$ & $45$\\
&                      &$1e-15$ & $\times$ & $\times$ & $953$ & $3818$ & $910$ & $871$ & $659$  & $\times$\\ [1ex]  \\ [-1.5ex]
&\multirow{2}{*}{$f_2$}&$1e-10$ & $4277(\textbf{62\%})$ & $\times$ & $\times$ & $\times$ & $16$ & $2095$ & $\times$ & $\times$\\
&                      &$1e-15$ & $4756(\textbf{42\%})$ & $\times$ & $\times$ & $\times$ & $50$ & $\times$ & $\times$ &$\times$\\ [1ex]  \\ [-1.5ex]
&\multirow{2}{*}{$f_3$}&$1e-10$ & $\times$ & $\times$ & $810(\textbf{98\%})$ &$\times$ & $534$ & $709$ & $385$ &$94$\\
&                      &$1e-15$ & $\times$ & $\times$ & $913(\textbf{98\%})$ & $\times$ & $913$ & $1111$ & $554$ & $\times$\\ [1ex]  \\ [-1.5ex]
\raisebox{4ex}[1.5pt] {$\begin{array}{c}
\mbox{Results}\\ \mbox{for} \\ \mbox{$60$-\emph{D}}
\end{array}$}&\multirow{2}{*}{$f_4$}&$1e-10$ & $1988(\textbf{92\%})$ & $2044$ & $318(\textbf{94\%})$ & $\times$ & $2$ & $2$ & $2$ & $2$\\
&                     &$1e-15$ & $2098(\textbf{28\%})$ & $2459(\textbf{56\%})$ & $356(\textbf{94\%})$ & $\times$ & $3$ & $3$ & $3$ & $3$\\ [1ex]  \\ [-1.5ex]
&\multirow{2}{*}{$f_5$}&$1e-10$ & $4239(\textbf{46\%})$ & $4970(\textbf{10\%})$ & $571(\textbf{94\%})$ & $1815$ & $25$ & $34$ & $254$ & $33$\\
&                      &$1e-15$ & $4814(\textbf{44\%})$ & $\times$ & $767(\textbf{94\%})$ & $2912$ & $72$ & $91$ & $562$ & $51$\\
\hline
\end{tabular}
\end{center}
%\vspace{-0.2cm}
\caption{Comparison of convergence speed  of PSO, DMS, CLPSO, HPSO and their variants.}
\label{CompCompComp}
\end{table*}

\subsection{A Modified HPSO Algorithm}

Next we take a look at the effect of the techniques on the HPSO algorithm \cite{Ratnaweera04}. The results are tabulated in Table~\ref{HPDE1}. We note that the proposed techniques result in improved performance for all test cases. Specifically for the Rosenbrock function, we note that all the proposed variants greatly improve the mean performance of the HPSO algorithm.

Although the results of the proposed D and DE variants are very promising, there is one important aspect in which the proposed DE scheme does not seem to be effective. While the combined effect of the proposed DE scheme has significantly reduced the number of computations for other PSO variants compared with the event-triggering cases, the result is the opposite in case of HPSO. As shown in Table~\ref{HPDE1}, the computational cost of the HPSO-DE algorithm is about $90\%$ of the HPSO algorithm for all functions. The reason behind this is the behavior of the HPSO algorithm. If the velocity of any of the particles goes to zero, the algorithm re-initializes the particle so that it keeps moving until the experiment ends. %Clearly, this is opposite to the purpose of the proposed algorithm.
Therefore, we go one step further in the case of HPSO and propose that the random re-initialization step be performed only if the particle has stopped of its own accord rather than being forced to stop by our techniques. We call this as the HPSO-DE2 variant. As a result of this extra step, we achieve remarkable results for HPSO. %These are shown in the last columns of Tables \ref{HPDE1} and \ref{HPDE2}.
The results show that for the HPSO-DE2 algorithm, the computations reduce to between $1\% - 20\%$ of the HPSO algorithm whereas for the HPSO-DE the computations are in the range of $90\%$. These results showcase the remarkable potential of the proposed techniques.

\subsection{Comparison of Convergence Times}

%For time-critical applications, the time required by an algorithm to convergence to a given error threshold is of particular importance.

The time required by an algorithm to reach a specified error threshold is of particular interest for time-critical and power-limited applications. Here we present a comparative analysis of the speed of convergence of the various algorithms to a specified error threshold. We compare the standard PSO, DMSPSO, CLPSO and HPSO along with their respective DE variants. Note that we use the HPOS-DE2 variant as it is more computationally efficient. We present results for two error thresholds, $1e-10$ and $1e-15$. Table~\ref{CompCompComp} shows the number of iterations needed by these algorithms to reach the specified error thresholds. These results are averaged over $50$ runs. For algorithms that do not reach the specified error threshold for all $50$ runs, the unsuccessful runs are excluded when calculating the convergence speed. For such cases, the value in bracket indicates the percentage of successful runs. An '$\times$' indicates that the algorithm failed to meet the error threshold for all $50$ runs.

The results show that not only do the DE variants converge much faster than the original algorithms, they also improve the success rate (PSO and PSO-DE for $30$ and $60$ dimensions cases, CLPSO and CLPSO-DE for $60$ dimensions case). Moreover, in some instances the original algorithm fails while its DE variant successfully converges to a solution, e.g., PSO-DE and DMS-DE for $f_2$. For $f_4$, the DE variants are able to converge in a remarkably fast time, i.e., $2-3$ iterations.

CLPSO is faster as compared to the standard PSO algorithm and generally has better performance. The modified PSO-DE not only outperforms the standard PSO but even the CLPSO algorithm in terms of convergence time to reach the set thresholds as shown in Table \ref{CompCompComp}. The results show that the proposed techniques improved the convergence times of all the PSO algorithms tested in this study, thereby increasing their scope for time critical applications.

\begin{table*}[t]
{\small
\hfill{}
\begin{center}
\begin{tabular}{c c c c c c c c c c c}
\hline
\textbf{f(x)} & \multicolumn{3}{|c|}{$f_1$} & \multicolumn{3}{c|}{$f_3$} &  \multicolumn{3}{c}{$f_5$} &\\
Values & \multicolumn{1}{|c}{Mean} & Iters. & \multicolumn{1}{c|}{Comp.} & Mean & Iters. & \multicolumn{1}{c|}{Comp.} & Mean & Iters. & Comp.\\
\hline \\ [-1ex]
{\bf PSO} & $7.1e\hspace{-.2em}-\hspace{-.2em}2$ & $500$ & $100\%$& $55.44$ & $500$ & $100\%$ & $5.62e\hspace{-.2em}-\hspace{-.2em}8 (\textbf{88\%})$ & $500$ & $100\%$\\
{\bf PSO-D} & $5.70e\hspace{-.2em}-\hspace{-.2em}53$ & $500$& $100\%$ & $0$ & $264$& $100\%$ & $7.18e\hspace{-.2em}-\hspace{-.2em}86$ & $500$ & $100\%$\\
{\bf PSO-DE} & $6.35e\hspace{-.2em}-\hspace{-.2em}20$ & $412$ & $70.33\%$ & $0$ & $293$ & $66.90\%$& $4.32e\hspace{-.2em}-\hspace{-.2em}21$ & $380$ & $67.76\%$\\
\\ [-1.5ex]
{\bf DMS} & $0.71 (\textbf{2\%})$ & $500$ & $100\%$ & $37.97$ & $500$ & $100\%$ & $2.28e\hspace{-.2em}-\hspace{-.2em}4$ & $500$ & $100\%$\\
{\bf DMS-D} & $3.64e\hspace{-.2em}-\hspace{-.2em}54$ & $500$ & $100\%$ & $0$ & $273$ & $100\%$ & $3.07e\hspace{-.2em}-\hspace{-.2em}61$ & $490$ & $100\%$\\
{\bf DMS-DE} & $3.85e\hspace{-.2em}-\hspace{-.2em}20$ & $392$ & $69.88\%$ & $0$ & $320$ & $66.48\%$ & $1.81e\hspace{-.2em}-\hspace{-.2em}21$ & $274$ & $67.40\%$ \\
\\ [-1.5ex]
{\bf CL} & $1.05e\hspace{-.2em}-\hspace{-.2em}47$ & $500$ & $60\%$ & $0$ & $312$ & $60\%$ & 1.90 $e\hspace{-.2em}-\hspace{-.2em}51$ & $500$ & $60\%$\\
{\bf CL-D} & $3.48e\hspace{-.2em}-\hspace{-.2em}51$ & $500$ & $60\%$ &  $0$ & $279$ & $60\%$ & $1.25e\hspace{-.2em}-\hspace{-.2em}53$ & $500$ & $60\%$\\
{\bf CL-DE} & $7.19e\hspace{-.2em}-\hspace{-.2em}19$ & $319$ & $39.24\%$ & $0$ & $258$ & $37.63\%$ & $2.09e\hspace{-.2em}-\hspace{-.2em}20$ & $299$ & $37.99\%$\\
\\ [-1.5ex]
{\bf HP} & $1.048e\hspace{-.2em}-\hspace{-.2em}5$ & $500$ & $80\%$ & $29.56$ & $500$ & $80\%$ & $5.78e\hspace{-.2em}-\hspace{-.2em}12$ & $500$ & $80\%$\\
{\bf HP-D} & $4.90e\hspace{-.2em}-\hspace{-.2em}111$ & $500$ & $80\%$ & $0$ & $87$ & $80\%$ & $1.63e\hspace{-.2em}-\hspace{-.2em}122$ & $500$ & $80\%$\\
{\bf HP-DE2} & $5.21e\hspace{-.2em}-\hspace{-.2em}15$ & $55$ & $3.99\%$ & $1.18e\hspace{-.2em}-\hspace{-.2em}13$ & $227$ & $12.80\%$ & $1.85e\hspace{-.2em}-\hspace{-.2em}16$ & $53$ & $3.62\%$\\
\hline
\end{tabular}
\end{center}}
\hfill{}
%\vspace{-0.2cm}
\caption{Comparison for 500 iterations (30-D case)}
\label{500DE1}
\end{table*}

\begin{table*}[t]
{\small
\hfill{}
\begin{center}
\begin{tabular}{c c c c c c c c c c c c c}
\hline
\textbf{f(x)} & \multicolumn{3}{|c|}{$f_1$} & \multicolumn{3}{c|}{$f_3$} &  \multicolumn{3}{c}{$f_5$} &\\
Values & \multicolumn{1}{|c}{Mean} & Iters. & \multicolumn{1}{c|}{Comp.} & Mean & Iters. & \multicolumn{1}{c|}{Comp.} & Mean & Iters. & Comp.\\
\hline \\ [-1ex]
{\bf PSO} & $\times$ & $\times$ & $\times$ & $150.12 (\textbf{98\%})$ & $500$ & $100\%$ & $\times$ & $\times$ & $\times$\\
{\bf PSO-D} & $1.26e\hspace{-.2em}-\hspace{-.2em}53$ & $500$ & $100\%$ & $0$ & $273$ & $100\%$ & $1e\hspace{-.2em}-\hspace{-.2em}57$ & $500$ & $100\%$\\
{\bf PSO-DE} & $1.46e\hspace{-.2em}-\hspace{-.2em}19$ & $446$ & $70.29\%$ & $0$ & $291$ & $66.94\%$ & $1e\hspace{-.2em}-\hspace{-.2em}21$ & $332$ & $73.67\%$\\
\\ [-1.5ex]
{\bf DMS} & $\times$ & $\times$ & $\times$ & $164.86 (\textbf{62\%})$ & $500$ & $100\%$ & $\times$ & $\times$ & $\times$\\
{\bf DMS-D} & $1.61e\hspace{-.2em}-\hspace{-.2em}54$ & $500$ & $100\%$ & $0$ & $269$ & $100\%$ & $2.09e\hspace{-.2em}-\hspace{-.2em}59$ & $492$ & $100\%$\\
{\bf DMS-DE} & $9.58e\hspace{-.2em}-\hspace{-.2em}20$ & $420$ & $69.90\%$ & $0$ & $304$ & $66.49\%$ & $2.16e\hspace{-.2em}-\hspace{-.2em}21$ & $344$ & $74.21\%$\\
\\ [-1.5ex]
{\bf CL} & $6.54e\hspace{-.2em}-\hspace{-.2em}44$ & $500$ & $60\%$ & $1.15$ & $500$ & $60\%$ & $5.63e\hspace{-.2em}-\hspace{-.2em}51 (\textbf{92\%})$ & $500$ & $60\%$\\
{\bf CL-D} & $2.57e\hspace{-.2em}-\hspace{-.2em}52$ & $500$ & $60\%$ & $0$ & $275$ & $60\%$ & $5.63e\hspace{-.2em}-\hspace{-.2em}52$ & $500$ & $60\%$\\
{\bf CL-DE} & $1.42e\hspace{-.2em}-\hspace{-.2em}18$ & $310$ & $39.19\%$ & $0$ & $259$ & $37.59\%$ & $1.68e\hspace{-.2em}-\hspace{-.2em}20$ & $300$ & $37.30\%$\\
\\ [-1.5ex]
{\bf HP} & $0.16$ & $500$ & $80\%$ & $69.05$ & $500$ & $80\%$ & $0.04 (\textbf{98\%})$ & $500$ & $80\%$ \\
{\bf HP-D} & $6.56e\hspace{-.2em}-\hspace{-.2em}106$ & $500$ & $80\%$ & $0$ & $92$ & $80\%$ & $6.52e\hspace{-.2em}-\hspace{-.2em}115$ & $500$ & $80\%$\\
{\bf HP-DE2} & $9.94e\hspace{-.2em}-\hspace{-.2em}15$ & $70$ & $3.98\%$ & $1.52e\hspace{-.2em}-\hspace{-.2em}13$ & $228$ & $12.78\%$ & $1.65e\hspace{-.2em}-\hspace{-.2em}16$ & $53$ & $16.40\%$\\
\hline
\end{tabular}
\end{center}}
\hfill{}
%\vspace{-0.2cm}
\caption{Comparison for 500 iterations (60-D case)}
\label{500DE2}
\end{table*}

\subsection{Robustness to Dimensionality}

\begin{figure}[t]
\centering  { { {
\includegraphics[width=.5\textwidth, clip=true, trim = 70 50 20 20 mm]{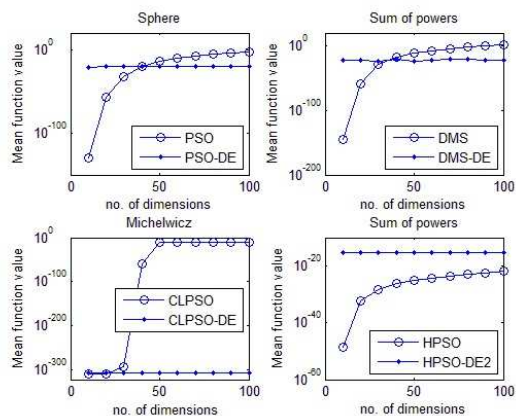}}}}
\caption[] {Effect of increasing dimensions on the mean performance of a few exemplary  DE variants.}\label{RobDim}
\end{figure}

Figure \ref{RobDim} shows the mean performance of various algorithms compared with their DE versions against the number of dimensions of the optimization problem. The figure shows that the mean performance of the DE technique is very robust to the increase in the number of dimensions of the problem. This is a very unique feature specific to the proposed D and DE techniques as nearly all PSO variants in literature suffer from the curse of dimensionality. Intuitively, this result makes sense as the dimension-wise approach takes care of the ``two step forward, one step backward'' problem and allows the particles to retain the best dimensions of a particular solution.

\subsection{A Comparative Study}

%Table \ref{CompStudyMeanPer} compares the performance of the original and the DE variants of the algorithms discussed above. For a fair comparison, the computational complexity of all algorithms  is given relative to that of the standard PSO. The results reaffirm that the DE variants of all algorithms converge faster, require significantly less computations and at the same time have good mean error performance. The only exception here is the CLPSO-DE algorithm which does not perform well for $f_2$. Furthermore, the proposed techniques have enhanced the performance of the basic PSO to a level similar to that of the other PSO variants.

A slightly different experiment is performed with the aim to test the performance of different PSO algorithms against their proposed D and DE variants in a scenario where only limited time is available. Instead of running the simulation for $5000$ iterations, the number is reduced to only $500$. The results for $3$ functions ($f_1$, $f_3$, $f_5$) are tabulated in Tables \ref{500DE1} and \ref{500DE2}. %The Rosenbrock function has been omitted because CLPSO performs poorly for this function so a fair comparison would not have been possible.
The Rosenbrock function has been omitted from this study because the original versions of all tested algorithms had a considerable higher mean performance as compared to their D and DE variants (except for CLPSO where even the D and DE variants performed poorly).  All algorithms performs almost equally well for the Michalewicz function so the results for this function have also been omitted. As can be seen from Tables \ref{500DE1} and \ref{500DE2}, almost all algorithms perform poorly for the $3$ functions when in their original form. For the $30$-dimension case, the D and DE variants of all algorithms outperform their original counterparts in terms of convergence speed, computational complexity and mean error performance. Only the CLPSO algorithm has a better mean error performance than its D and DE variants but it still takes longer to converge and has a higher computational cost. For the $60$-dimension case, PSO, DMS and HPSO fail to converge for all three test functions while CLPSO fails for $f_3$. In contrast, their D and DE variants are able to converge in all cases. Application of the proposed techniques results in significant performance improvement with considerable reduction in computations. This experiment shows the prowess of the proposed techniques, particularly in time-critical situations.

\section{Conclusion}
\label{Conclusion}

The paper proposes two techniques for the PSO family of algorithms. The first technique reduces computations while the second technique increases speed of convergence. The event-triggering approach sacrifices performance for reduction in computational complexity. The strategy is to calculate the cognitive term of the update equation for the $d$-th dimension of a particle only if
the distance between this dimension of the particle's current position and its local best is larger than a predefined threshold. Similarly, the social term is calculated only if the distance between the particle's current position in the $d$-th dimension and the $d$-th dimension of its global best exceeds a threshold. The dimension-wise technique allows the PSO algorithm to learn from the rich information contained in the various dimensions of the current positions, and local best positions of its particles. The technique works for separable functions and minimizes each term of the cost function independently resulting in a much faster convergence. This does not incur any extra computations as compared to the standard function evaluation procedure. The proposed techniques can be applied separately or in conjunction to other PSO algorithms presented in literature. Simulation results show that the application of the proposed techniques to PSO variants improve their convergence speeds and reduce their computational complexities while still maintaining an acceptable error performance; features that are desirable for time-critical applications as well as those with power constraints.

\end{document}